\documentclass[conference]{IEEEtran}
\usepackage[cmex10]{amsmath}
\usepackage{url}
\usepackage[utf8]{vietnam}
\usepackage{float}
\floatstyle{plaintop}
\restylefloat{table}
\usepackage{tikz-qtree}
\usepackage{graphicx}
\usepackage{caption}
\usepackage{subcaption}
\usepackage{fixltx2e}
\usepackage{float}
\usepackage{tikz}
\usepackage{tikz-qtree, tikz-qtree-compat}
\usetikzlibrary{fit,shapes}
\usetikzlibrary{plotmarks}
\usepackage[ruled,vlined]{algorithm2e}
\usepackage{tikz-dependency}
\usepackage{mathtools}
\usepackage[utf8]{inputenc}

\begin{document}
\restylefloat{table}
\restylefloat{figure}

\captionsenglish
%


\title{Building a Semantic Role Labelling System for Vietnamese}

 \author{\IEEEauthorblockN{Thai-Hoang Pham}
   \IEEEauthorblockA{FPT University}
   \textit{hoangpt@fpt.edu.vn}
 \and
   \IEEEauthorblockN{Xuan-Khoai Pham}
   \IEEEauthorblockA{FPT University}
   \textit{khoaipxse02933@fpt.edu.vn}
 \and
   \IEEEauthorblockN{Phuong Le-Hong}
   \IEEEauthorblockA{Hanoi University of Science}
   \textit{phuonglh@vnu.edu.vn}
 }


\maketitle
\begin{abstract}
  Semantic role labelling (SRL) is a task in natural language
  processing which detects and classifies the semantic arguments
  associated with the predicates of a sentence. It is an important
  step towards understanding the meaning of a natural language. There
  exists SRL systems for well-studied languages like English, Chinese
  or Japanese but there is not any such system for the Vietnamese
  language. In this paper, we present the first SRL system for
  Vietnamese with encouraging accuracy. We first demonstrate that a
  simple application of SRL techniques developed for English could not
  give a good accuracy for Vietnamese. We then introduce a new
  algorithm for extracting candidate syntactic constituents, which is
  much more accurate than the common node-mapping algorithm usually
  used in the identification step. Finally, in the classification step, in
  addition to the common linguistic features, we propose novel and
  useful features for use in SRL. Our SRL system achieves an $F_1$
  score of 73.53\% on the Vietnamese PropBank corpus. This system, including
  software and corpus, is available as an open source project and we
  believe that it is a good baseline for the development of future
  Vietnamese SRL systems.
\end{abstract}

\IEEEpeerreviewmaketitle

\section{Introduction}
SRL is the task of identifying semantic roles of predicates in the
sentence. In particular, it answers a question \textit{Who did What to
  Whom, When, Where, Why?}. A simple Vietnamese sentence \textit{Nam
  giúp Huy học bài vào hôm qua} (Nam helped Huy to do homework
yesterday) is given in 
Figure~\ref{fig:1}.
\begin{figure}[h!]
\[
\underbracket[0.5px]{\text{Nam}}_{Who} \; gi\acute{u}p \; \underbracket[0.5px]{\text{Huy}}_{Whom} \; \underbracket[0.5px]{\text{học bài}}_{What} \; \underbracket[0.5px]{\text{v\`{a}o hôm qua}}_{When}
\]
\caption{An example sentence}\label{fig:1}
\end{figure}

To assign semantic roles for the sentence above, we must analyse and
label the propositions concerning the predicate \textit{giúp} (helped) of the
sentence. Figure~\ref{fig:2} shows a result of the SRL for this
example, where meaning of the labels will be described in detail in
Section~\ref{sec:exp}.

\begin{figure}[h!]
\[
\underbracket[0.5px]{\text{Nam}}_{Arg0} \; gi\acute{u}p \; \underbracket[0.5px]{\text{Huy}}_{Arg1} \; \underbracket[0.5px]{\text{học bài}}_{Arg2} \; \underbracket[0.5px]{\text{v\`{a}o hôm qua}}_{ArgM-TMP}
\]
\caption{Semantic roles for the example sentence} \label{fig:2}
\end{figure}

SRL has been used in many natural language processing (NLP)
applications such as question answering~\cite{Shen:2007}, machine
translation~\cite{Lo:2010}, document summarization~\cite{Aksoy:2009} and information
extraction~\cite{Christensen:2010}. Therefore, SRL is an important task in NLP.

The first SRL system was developed by Gildea and
Jurafsky~\cite{Gildea:2002}. This system was performed on the FrameNet
corpus and was used for English. After that, SRL task has been widely
researched by the NLP community. In particular, there have been two
shared-tasks, CoNLL-2004~\cite{Carreras:2004} and
CoNLL-2005~\cite{Carreras:2005}, focusing on SRL task for English. Most
of the systems participating in these share-tasks treated this problem
as a classification problem and applied some supervised machine
learning techniques. In addition, there were some systems developed
for other languages such as Chinese~\cite{Xue:2005} or Japanese~\cite{Tagami:2009}.

In this paper, we present the first SRL system for Vietnamese with
encouraging accuracy. We first demonstrate that a simple application
of SRL techniques developed for English or other languages could not
give a good accuracy for Vietnamese. In particular, in the constituent
identification step, the widely used 1-1 node-mapping algorithm for
extracting argument candidates performs poorly on the Vietnamese
dataset, having $F_1$ score of 35.84\%. We thus introduce a new
algorithm for extracting candidates, which is much more accurate,
achieving an $F_1$ score of 83.63\%.

In the classification step, in addition to the common linguistic
features, we propose novel and useful features for use in SRL,
including function tags and word clusters obtained by performing a
Gaussian mixture analysis on the distributed representations of
Vietnamese words. These features are employed in two statistical
classification models, Maximum Entropy and Support Vector
Machines, which are proved to be good at many classification
problems. 

Our SRL system achieves an $F_1$ score of 73.53\% on the Vietnamese PropBank
corpus. This system, including software and corpus, is available as an
open source project and we believe that it is a good baseline for the
development of future Vietnamese SRL systems.

The paper is structured as follows. Section~\ref{sec:background}
introduces briefly the SRL task and two well-known corpora for
English. Section~\ref{sec:method} describes the methodologies of some
existing systems and of our system. Some difficulties of SRL for
Vietnamese are also discussed. Section~\ref{sec:exp} presents the
evaluation results and discussion. Finally,
Section~\ref{sec:conclusion} concludes the paper and suggests some
directions for future work.

\section{Background}\label{sec:background}

\subsection{SRL Task Description}
The SRL task is usually divided into two steps. The first step is
argument identification. The goal of this step is to identify the
syntactic constituents of a sentence which are the most likely to be
semantic arguments of its predicates. This is a difficult problem
since the number of constituent candidates is exponentially large,
especially for long sentences.

The second step is argument classification which decides the exact
semantic role for each constituent candidate identified in the first
task. For example, the identification step of the sentence in the
previous example \textit{Nam giúp Huy học bài vào hôm qua} is described
in Figure~\ref{fig:3} and in the classification task, semantic roles
are labelled as shown Figure~\ref{fig:2}.
\begin{figure}[t] 
\[
\underbracket[0.5px]{\text{Nam}} \; gi\acute{u}p \;
\underbracket[0.5px]{\text{Huy}} \; \underbracket[0.5px]{\text{học bài}} \; \underbracket[0.5px]{\text{v\`{a}o hôm qua}} 
\]
\caption{Example of identification task}\label{fig:3}
\end{figure}

\subsection{Existing Corpora for SRL}
\subsubsection{FrameNet}
The FrameNet project is a lexical database of English. It was built by
annotating examples of how words are used in actual texts. It consists
of more than 10,000 word senses, most of them with annotated examples
that show the meaning and usage and more than 170,000 manually
annotated sentences~\cite{Baker:2003}. This is the most widely used
dataset upon which SRL systems for English have been developed and
tested.

FrameNet is based on the Frame Semantics theory~\cite{Boas:2005}. The basic idea is that
the meanings of most words can be best understood on the basis of a
semantic frame: a description of a type of event, relation, or entity
and the participants in it. All members in semantic frames are called
frame elements. For example, a sentence in FrameNet is annotated in
cooking concept as shown in Figure~\ref{fig:4}.
\begin{figure}[h!] 
\[
\underbracket[0.5px]{\text{The boy}}_{Cook} \; grills \; \underbracket[0.5px]{\text{their catches}}_{Food} \; \underbracket[0.5px]{\text{on an open fire}}_{Heating-instrument} \; 
\]
\caption{An example sentence in the FrameNet corpus}\label{fig:4}
\end{figure}
\subsubsection{PropBank}
PropBank is a corpus that is annotated with verbal propositions and
their arguments~\cite{Babko:2005}. PropBank tries to supply a general purpose labelling
of semantic roles for a large corpus to support the training of
automatic semantic role labelling systems. However, defining such a
universal set of semantic roles for all types of predicates is a
difficult task; therefore, only Arg0 and Arg1 semantic roles can be
generalized. In addition to the core roles, PropBank defines several
adjunct roles that can apply to any verb. It is called Argument
Modifier. The semantic roles covered by the PropBank are the following:
\begin{itemize}
\item \textbf{Core Arguments} (Arg0-Arg5, ArgA): Arguments define
  predicate specific roles. Their semantics depend on predicates in
  the sentence.
\item \textbf{Adjunct Arguments} (ArgM-): General arguments that can 
  belong to any predicate. There are 13 types of adjuncts.
\item \textbf{Reference Arguments} (R-): Arguments represent arguments
  realized in other parts of the sentence.
\item \textbf{Predicate} (V): Participant realizing the verb of the
  proposition.
\end{itemize}
For example, the sentence of Figure~\ref{fig:4} can be annotated in 
the PropBank role schema as shown in Figure~\ref{fig:5}.
 
\begin{figure}[t] 
\[
\underbracket[0.5px]{\text{The boy}}_{Arg0} \; grills \;
\underbracket[0.5px]{\text{their catches}}_{Arg1} \;
\underbracket[0.5px]{\text{on an open fire}}_{Arg2} \;  
\]
\caption{An example sentence in the PropBank corpus} \label{fig:5}
\end{figure}

\section{Methodology}\label{sec:method}
\subsection{Existing Approaches}
This section summarizes existing approaches used by typical SRL systems
for well-studied languages. We describe these systems by investigating
two aspects, namely data type that the systems use and their
strategies for labelling semantic roles, including model types,
labelling strategies and degrees of granularity.

\subsubsection{Data Types}
There are some kinds of data used in the training of SRL systems. Some
systems use bracketed trees as the input data.  A bracketed tree of a
sentence is the tree of nested constituents representing its
constituency structure. Some systems use dependency trees of a
sentence, which represents dependencies between individual words of a
sentence.  The syntactic dependency represents the fact that the
presence of a word is licensed by another word which is its
governor. In a typed dependency analysis, grammatical labels are added
to the dependencies to mark their grammatical relations, for example
\textit{nominal subject} (nsubj) or \textit{direct object}
(dobj). Figure~\ref{fig:6} shows the bracketed tree and the
dependency tree of an example sentence.

\begin{figure}[h!]
\centering
\centering
\begin{tabular}{cc}
\begin{tikzpicture}
\tikzset {level distance=20pt}
\tikzset {frontier/.style={distance from root=80pt}}
\Tree [.S [.N \edge[dashed]; Nam ]
		  [.VP [.V \edge[dashed]; đá ]
		  	   [.NP [.N \edge[dashed]; bóng ] ] ] ]
\end{tikzpicture} &
\begin{dependency}
\begin{deptext}
Nam \& đá \& bóng \\
N   \& V  \& N \\
\end{deptext}
\deproot{2}{root}
\depedge{2}{1}{nsubj}
\depedge{2}{3}{dobj}
\end{dependency} \\
(a) {The bracketed tree} & (b) {The dependency tree}
\end{tabular}
\caption{Bracketed and dependency trees for sentence \textit{Nam đá
    bóng} (Nam plays football)} \label{fig:6}
\end{figure}

\subsubsection{SRL Strategy}
\paragraph{Model Types}
There are two types of classification models: Independent Model and
Joint Model. While independent model decides the label of each
argument's candidate independently of other candidates, joint model
finds the best overall labelling for all candidates in the
sentence. Independent model runs fast but are prone to
inconsistencies. For example, Figure~\ref{fig:7} shows some typical
inconsistencies, including overlapping arguments, repeating arguments
and missing arguments of a sentence \textit{Do học chăm, Nam đã đạt
  thành tích cao} (By studying hard, Nam got a high achievement). 

\begin{figure}[h!]
\centering
\begin{subfigure}{0.33 \textheight}
\centering
\[
\text{Do}\underbracket{\text{ học chăm}}_{Arg1}, \text{ Nam đã \textit{đạt} thành tích cao.}
\]
\[
\text{Do học} \underbracket{\text{chăm, Nam}}_{Arg1}\text{ đã \textit{đạt} thành tích cao.}
\]
\caption{Overlapping argument}
\end{subfigure}
\begin{subfigure}{0.33 \textheight}
\centering
\[
\text{Do} \underbracket{\text{học}}_{Arg1} \text{chăm}, \underbracket{\text{Nam}}_{Arg1} \text{đã \textit{đạt} thành tích cao.}
\]
\caption{Repeating argument}
\end{subfigure}
\begin{subfigure}{0.33 \textheight}
\centering
\[
\underbracket{\text{Do học chăm, Nam}}_{Arg0} \text{đã \textit{đạt}} \underbracket{\text{thành tích cao}}_{Arg0}.
\]
\caption{Missing argument}
\end{subfigure}
\caption{An example of inconsistencies}\label{fig:7}
\end{figure}

\paragraph{Labelling Stategies}
Strategies for labelling semantic roles are diverse, but we can
summarize that there are three main strategies. Most of the systems
use a two-step approach consisting of identification and
classification~\cite{Punyakanok:2005, Haghighi:2005}. The first step identifies arguments from
many candidates. It is essentially a binary classification
problem. The second step classifies these arguments
into particular semantic roles. Some systems use single
classification step by adding a ``null'' label into semantic roles,
denoting that this is not an argument~\cite{Surdeanu:2005}. Other systems consider SRL as a
sequence tagging~\cite{Marquez:2005, Pradhan:2005}.

\paragraph{Granularity}
Existing SRL systems use different degrees of granularity when
considering constituents. Some systems use individual words as their
input and perform sequence tagging to identify arguments. This
method is called Word-by-Word (W-by-W) approach. Other systems
directly take syntactic phrases as input constituents. This method is
called Constituent-by-Constituent (C-by-C) approach. 

Compared to the W-by-W approach, C-by-C approach has several 
advantages. First, phrase boundaries are usually consistent with
argument boundaries. Second, C-by-C approach allows us to work with
larger contexts due to a smaller number of candidates in comparison to
the W-by-W approach.

\subsection{Our Approach}

The previous subsection has reviewed existing techniques for SRL which
have been published so far for well-studied languages. In this
section, we first show that these techniques per se cannot give a good
result for Vietnamese SRL, due to some inherent difficulties, both in
terms of language characteristics and of the available corpus. We then
develop a new algorithm for extracting candidate constituents for use
in the identification step.

Some difficulties of Vietnamese SRL are related to its SRL corpus. We
use the Vietnamese PropBank~\cite{ViPropBank:2014} in the development
of our SRL system.\footnote{To our knowledge, this is the first SRL
  corpus for Vietnamese which has been published for free research.}
This SRL corpus has 5,000 annotated sentences, which is
much smaller than SRL corpora of other languages. For example, the
English PropBank contains about 50,000 sentences, which is ten times
larger. While smaller in size, the Vietnamese PropBank has
more semantic roles than the English PropBank has -- 25 roles compared
to 21 roles. This makes the unavoidable data sparseness problem more
severe for Vienamese SRL than for English SRL. 

In addition, our extensive inspection and experiments on the Vietnamese
PropBank have uncovered that this corpus has many annotation errors,
largely due to encoding problems and inconsistencies in annotation. In
many cases, we have to fix these annotation errors by ourselves. In
other cases where only a proposition of a complex sentence is
incorrectly annotated, we perform an automatic preprocessing procedure
to drop it out, leave the correctly annotated propositions
untouched. We finally come up with a corpus of 4,800 sentences which
are semantic role annotated. This corpus will be released for free use
for research purpose.

A major difficulty of Vietnamese SRL is due to the nature of the
language, where its linguistic characteristics are different from
occidental languages~\cite{Le-Hong:2015}. We first try to apply the
common node-mapping algorithm which are widely used in English SRL
systems to the Vietnamese corpus. However, this application gives us a
very poor performance. Therefore, in the identification step, we
develop a new algorithm for extracting candidate constituents which is
much more accurate for Vietnamese than the node-mapping
algorithm. Details of experimental results will be provided in the
Section~\ref{sec:exp}

In order to improve the accuracy of the classification step,
and hence of our SRL system as a whole, we have integrated many useful
features for use in two statistical classification models, namely
Maximum Entropy (ME) and Support Vector Machines (SVM). On the one
hand, we adapt the features which have been proved to be good for SRL
of English. On the other hand, we propose some novel features,
including function tags and word clusters.

In the next paragraph, we present our constituent extraction
algorithm for the identification step. Details of the features for use
in the classification step will be presented in Section~\ref{sec:exp}.

\subsubsection{Constituent Extraction Algorithm}

This algorithm aims to extract constituents from a bracketed
tree which are associated to their corresponding predicates of 
the sentence. If the sentence has multiple predicates, multiple
constituent sets corresponding to the predicates are extracted. Pseudo
code of the algorithm is described in Algorithm~\ref{alg:1}. 

\begin{algorithm}[t]
\DontPrintSemicolon
\SetKwInOut{Input}{input}\SetKwInOut{Output}{output}
\Input{A bracketed tree $T$ and its predicate}
\Output{A tree with constituents for the predicate}
\Begin{
$currentNode \leftarrow predicateNode$\;
\While{$currentNode \neq$ T.root()}{
	\For{$S \in currentNode$.sibling()}{
		\If{$|S.children()| > 1$ and $S.children().get(0).isPhrase()$}{
			$sameType \leftarrow true$\\
			$diffTag \leftarrow true$\\
			$phraseType \leftarrow S.children().get(0).phraseType()$\\
			$funcTag \leftarrow S.children().get(0).functionTag()$\\
			\For{$i\leftarrow 1$ \KwTo $|S.children()| - 1$}{
				\If{$S.children().get(i).phraseType() \neq phraseType$}{
					$sameType \leftarrow false$\\
					break
				}
				\If{$S.children().get(i).functionTag() = funcTag$}{
					$diffTag \leftarrow false$\\
					break
				}
			}
			\If{$sameType$ and $diffTag$}{
				\For{$child \in S.children()$}{
					$T.collect(child)$
				}
			}
		}
		\Else{
			$T.collect(S)$
		}
	}
	$currentNode \leftarrow currentNode.parent()$
}
return $T$
}
\caption{Constituent Extraction Algorithm}\label{alg:1}
\end{algorithm}

This algorithm uses several simple functions. The $root()$ function
gets the root of a tree. The $children()$ function gets the children
of a node. The $sibling()$ function gets the sisters of a node. The $isPhrase()$
function checks whether a node is of phrasal type or not. The
$phraseType()$ function and $functionTag()$ function extracts the phrase type 
and function tag of a node, respectively. Finally, the
$collect(node)$ function collects words from leaves of the subtree
rooted at a node and creates a constituent.

\begin{figure}[h!]
\begin{tiny}
\centering
\begin{subfigure}{0.2 \textheight}
\centering
\tikzset {level distance=15pt}
\begin{tikzpicture}
\Tree
[.S [.NP-SUB [.N-H Bà ] ]
	[.VP [.V-H nói ]
		 [.SBAR [.S [.NP-SUB [.P-H nó ] ]
		  	   	    [.VP [.\node[draw,ellipse]{V-H}; \textit{là} ]
		  	   			 [.NP [.N-H {con trai} ]
		  	   			 	  [.P tôi ]
		  	   			 	  [.T mà ] ] ] ] ] ] ] ]
\end{tikzpicture}
\end{subfigure}
\begin{subfigure}{0.2 \textheight}
\centering
\tikzset {level distance=15pt}
\begin{tikzpicture}
\Tree
[.S [.NP-SUB [.N-H Bà ] ]
	[.VP [.V-H nói ]
		 [.SBAR [.S [.NP-SUB [.P-H nó ] ]
		  	   	    [.VP [.\node[draw,ellipse]{V-H}; \textit{là} ]
		  	   			 [.NP \edge[roof]; {con trai tôi mà } ] ] ] ] ] ]
\end{tikzpicture}
\end{subfigure}
\begin{subfigure}{0.2 \textheight}
\centering
\tikzset {level distance=15pt}
\begin{tikzpicture}
\Tree
[.S [.NP-SUB \edge[roof]; {Bà} ]
	[.\node[draw, ellipse]{VP}; [.V-H \edge[roof]; nói ]
		 [.\node[draw, ellipse]{SBAR}; [.\node[draw, ellipse]{S}; [.NP-SUB \edge[roof]; {nó} ]
		  	   	    [.\node[draw, ellipse]{VP}; [.V-H \textit{là} ]
		  	   			 [.NP \edge[roof]; {con trai tôi mà } ] ] ] ] ] ]
\end{tikzpicture}
\end{subfigure}
\caption{Extracting constituents of the sentence ``Bà nói nó
    là con trai tôi mà'' at predicate ``\textit{là}''}
\label{fig:8}
\end{tiny}
\end{figure}
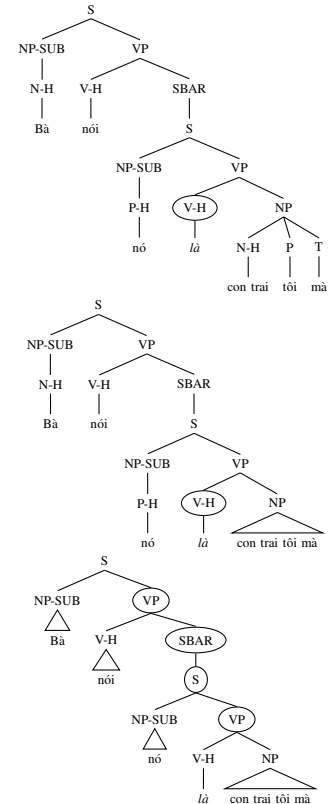

Figure~\ref{fig:8} shows an example of running the algorithm on a
sentence \textit{Bà nói nó là con trai tôi mà} (She said that he is my
son). First, 
we find the current predicate node V-H \textit{là} (is). The current
node has only one sibling NP. This node has one 
child, so its associated words are collected. After that, we set current node to
its parent and repeat the process until reaching the root
of the tree. Finally, we obtain a tree with constituents: \textit{Bà},
\textit{nói}, \textit{nó}, and \textit{con trai tôi mà}. 

\subsubsection{Our SRL System}

Our SRL system is developed on the Vietnamese PropBank. It thus operates on
fully bracketed trees. We employ ME and SVM as classifiers. Its
classification model is of type independent and its input are C-by-C.

\section{Experiment}\label{sec:exp}

In this section, we first introduce the Vietnamese PropBank upon which
our SRL system has been trained and tested. We then propose two
feature sets in use. Finally, we present and discuss experimental results.

\subsection{Dataset}

We conduct experiments on the Vietnamese
PropBank~\cite{ViPropBank:2014} containing about 5,460 sentences which
are manually annotated with semantic roles. This corpus has a similar
annotation schema to the English PropBank. Due to some inconsistency
annotation errors of the corpus, notably in many complex sentences, we
were not able to use all the corpus in our experiments. We focus
ourselves in simple sentences which have only one predicate rather
than complex sentences with multiple predicates. After extracting
sentences, we have a corpus of about 4,860 simple sentences
which are annotated with semantic roles.

The semantic roles covered by the Vietnamese PropBank are the
following:
\begin{itemize}
\item \textbf{Core Arguments} (Arg0-Arg4): Arguments define predicate
  specific roles. These core arguments are similar to those of the
  English PropBank, however, there are 5 roles instead of 7, compared
  to the English PropBank.
\item \textbf{Adjunct Arguments} (ArgM-): There are 20 types of
  adjuncts, as listed in Table~\ref{tab:adjuncts}.
\item \textbf{Predicate} (V): In Vietnamese, a predicate is not only a
  verb, but it could be also a noun, an adjective or a preposition.
\end{itemize}

\begin{table}[t]
\caption{Adjunct arguments in Vietnamese}
\centering
\begin{scriptsize}
\begin{tabular}{|l|l||l|l|}
\hline
Role Name&Description&Role Name&Description\\
\hline
ArgM-ADV & general-purpose & ArgM-CAU & cause \\ 
ArgM-DIS & discourse marker & ArgM-DIR & direction \\ 
ArgM-NEG & negation marker & ArgM-MNR & manner \\ 
ArgM-PRD & predication & ArgM-PRP & purpose \\ 
ArgM-MOD & modal verb & ArgM-TMP & temporal \\ 
ArgM-REC & reciprocal & ArgM-GOL & goal \\ 
ArgM-LVB & light verb & ArgM-EXT & extent \\ 
ArgM-COM & comitative & ArgM-I & interjection \\ 
ArgM-Partice & partice & ArgM-PNC & purpose \\ 
ArgM-ADJ & unknown & ArgM-RES & unknown \\ 
\hline
\end{tabular}
\end{scriptsize}
 \label{tab:adjuncts}
\end{table}

\subsection{Feature Sets}

We use two feature sets in this study. The first one is composed of
basic features which are commonly used in SRL system for English. This
feature set is used in the SRL system of Gildea and Jurafsky on the
FrameNet corpus~\cite{Gildea:2002}. 

\subsubsection{Basic Feature Set}

This feature set consists of 6 feature templates, as follows:
\begin{enumerate}
\item \textbf{Phrase Type}: This is very useful feature in classifying semantic roles
because different roles tend to have different syntactic
categories. For example, in the sentence in Figure~\ref{fig:8} \textit{Bà nói nó
  là con trai tôi mà}, the phrase type of constituent \textit{nó} is
\textit{NP}. 
\item \textbf{Parse Tree Path}: This feature captures the syntactic relation between a
constituent and a predicate in a bracketed tree. This is the
shortest path from a constituent node to a predicate node in the
tree. We use either symbol $\uparrow$ or symbol $\downarrow$ to indicate the upward
direction or the downward direction, respectively. For example, the
parse tree path from constituent \textit{nó} to the predicate
\textit{l\`{a}} is \textit{NP$\uparrow$S$\downarrow$VP$\downarrow$V}.
\item \textbf{Position}: Position is a binary feature that describes whether
  the constituent occurs after or before the predicate. 
  It takes value \textit{0} if the constituent appears
  before the predicate in the sentence or value \textit{1}
  otherwise. For example, the position of constituent \textit{nó} in
  Figure~\ref{fig:8} is \textit{0} since it appears before predicate
  \textit{là}.
\item \textbf{Voice}: Sometimes, the differentiation between active and passive voice is
useful. For example, in an active sentence, the subject is
usually an \textit{Arg0} while in a passive sentence, it is often an
\textit{Arg1}. Voice feature is also binary feature, taking value \textit{1} for
active voice or \textit{0} for passive voice. The sentence in
Figure~\ref{fig:8} is of active voice, thus its voice feature value is \textit{1}.
\item \textbf{Head Word}: This is the first word of a phrase. For
  example, the head word for the phrase \textit{con trai tôi mà} is
  \textit{con trai}. 
\item \textbf{Subcategorization}: Subcategorization feature captures the tree that
has the concerned predicate as its child. For example, in
Figure~\ref{fig:8}, the subcategorization of the predicate
\textit{l\`{a}} is \textit{VP(V, NP)}.
\end{enumerate}

\subsubsection{Modified Features and New Features}

Preliminary investigations on the basic feature set give us a rather poor
result. Therefore, we propose some modified features and novel
features so as to improve the accuracy of the system. These 
features are as follows:
\begin{enumerate}
\item  \textbf{Function Tag}: Function tag is a useful information,
especially for classifying adjunct arguments. It determines a
constituent's role, for example,  the function tag of constituent
\textit{nó} is \textit{SUB}, indicating that this has a subjective role.
\item \textbf{Partial Parse Tree Path}:
Many sentences have complicated structure. It can make parse tree path
very long and infrequent. We propose to cut a
path from the lowest common ancestor to its predicate, instead of
using the full path. For example, the partial path from the constituent \textit{nó} to the predicate \textit{là} in
Figure~\ref{fig:8} is \textit{NP$\uparrow$S}.  
\item \textbf{Distance}: This feature records the length of the full
  parse tree path before pruning. This feature helps retaining some
  information that might be lost when a partial path, instead of a
  full path, is used. For example, the distance from constituent
\textit{nó} to the predicate \textit{là} is \textit{3}. 
\item \textbf{Predicate Type}: Unlike in English, the type of predicates
  in Vietnamese is much more complicated. It is not only a verb, but is also a noun, an
  adjective, or a preposition. Therefore, we propose
a new feature which captures predicate types. For example, the
predicate type of the concerned predicate is \textit{V}. 
\item \textbf{Word Cluster}: Word clusters have been shown to help
  improve the performance of many NLP tasks because they alleviate the
  severity of the data sparseness problem. Thus, in this work we
  propose to use word cluster 
  features. We first produce distributed word representations (or word
  embeddings) of Vietnamese words, where each word is represented by a
  dense, real-valued vector of 50 dimensions, by using a Skip-gram
  model described in~\cite{Mikolov:2013a,Mikolov:2013b}. We then
  cluster these word vectors into 128 groups using a Gaussian mixture
  model.\footnote{Actually, there is an additional group for unknown
    words.} A word cluster feature is defined as the cluster identifier
  of the concerned word.
\end{enumerate}

\subsection{Results and Discussions}

\subsubsection{Evaluation Method}

We use a 10-fold cross-validation method to evaluate our system. The
final accuracy scores is the average scores of the 10 runs.

The evaluation metrics are the precision, recall and $F_1$-measure. The
precision ($P$) is the proportion of labelled arguments identified by
the system which are correct; the recall ($R$) is the proportion of
labelled arguments in the gold results which are correctly identified
by the system; and the $F_1$-measure is the harmonic mean of $P$ and
$R$, that is $F_{1} = 2PR/(P+R)$.

\subsubsection{Baseline System}

In the first experiment, we compare our constituent extraction
algorithm to the 1-1 node mapping algorithm. Table~\ref{tab:per1}
shows the performance of two extraction algorithms.

\begin{table}
\centering
\begin{tabular}{|l|c|c|}
\hline 
 & 1-1 Node Mapping & Our Extraction Alg. \\ 
\hline 
Precision & 29.53\% & 81.00\% \\ 
\hline 
Recall & 45.60\% & 86.43\% \\ 
\hline 
F1 & 35.84\% & 83.63\% \\ 
\hline 
\end{tabular} 
\caption{Accuracy of two extraction algorithms}\label{tab:per1}
\end{table}

We see that our extraction algorithm outperforms significantly the 1-1
node mapping algorithm, in both of the precision and the recall
ratios. In particular, the precision of the 1-1 node mapping algorithm
is only 29.53\%; this means that this method captures many candidates
which are not arguments. In contrast, our algorithm is able to
identify a large number of correct argument candidates, particularly
with the recall ratio of 86.43\%. This result clearly demonstrates
that we cannot take for granted that a good algorithm for
English could also work well for another language of different
characteristics. 

In the second experiment, we continue to compare the performance of
the two extraction algorithms, this time at the final classification
step and get the baseline for Vietnamese SRL. The classifier we use
in this experiment is a Maximum Entropy classifier.\footnote{We use the logistic regression classifier with $L_2$
  regularization provided by the \texttt{scikit-learn} software
  package. The regularization term is fixed at 1.} Table~\ref{tab:per2} shows the accuracy of the baseline system.

\begin{table}[h!]
\centering
\begin{tabular}{|l|c|c|}
\hline 
 & 1-1 node mapping & Our Extraction Alg. \\ 
\hline 
Precision & 52.80\% & 53.79\% \\ 
\hline 
Recall & 3.30\% & 47.51\% \\ 
\hline 
F1 & 6.20\% & 50.45\% \\ 
\hline 
\end{tabular} 
\caption{Accuracy of baseline system} \label{tab:per2}
\end{table}

One again, this result confirms that our algorithm is much superior
than the 1-1 node mapping algorithm. The $F_1$ of our baseline SRL
system is 50.45\%, compared to 6.20\% of the 1-1 node mapping
system. This result can be explained by the fact that the 1-1 node
mapping algorithm has a very low recall ratio, because it
identifies incorrectly many argument candidates.

\subsubsection{Labelling Strategy}
In the third experiment, we compare two labelling strategies for
Vietnamese SRL (cf. Section~\ref{sec:method}). In addition to the ME
classifier, we also try the Support Vector Machine (SVM) classifier,
which usually gives good accuracy in a wide variety of classification
problems.\footnote{We use a linear SVM provided in the
  \texttt{scikit-learn} software package with default parameter
  values.} Table~\ref{tab:per3} shows the $F_1$ scores of different
labelling strategies.

\begin{table}[h!]
\centering
\begin{tabular}{|l|c|c|}
 \hline 
  & ME & SVM \\ 
 \hline 
 1-step strategy & 50.45\% & 68.91\%  \\ 
 \hline 
 2-step strategy & 49.76\% & 68.55\% \\ 
 \hline 
 \end{tabular}  
\caption{Accuracy of two labelling strategies}\label{tab:per3}
\end{table}

We see that the SVM classifier outperforms ME the classifier by a
large margin. The best accuracy is obtained by using 1-step stragegy
with SVM classifier. The current SRL system achieves an $F_1$ score of
68.91\%.

\subsubsection{Feature Analysis}

In the fourth experiment, we analyse and evaluate the impact of each
individual feature to the accuracy of our system so as to find the
best feature set for our Vietnamese SRL system. We start with the basic
feature set presented previously, denoted by $\Phi_0$ and augment it
with modified and new features as shown in Table~\ref{tab:fs-1}. The
accuracy of these feature sets are shown in Table~\ref{tab:per4-1}.

\begin{table}[h!]
\centering
\begin{tabular}{|c|l|}
\hline 
Feature Set & Description \\ 
\hline 
$\Phi_{1}$ & $\Phi_{0} \cup $\{Function Tag\} \\ 
\hline 
$\Phi_{2}$ & $\Phi_{0} \cup $\{Predicate Type\} \\ 
\hline 
$\Phi_{3}$ & $\Phi_{0 }\cup $\{Distance\} \\ 
\hline 
\end{tabular}
\caption{Feature sets}\label{tab:fs-1}
\end{table} 

\begin{table}[h!]
\centering
\begin{tabular}{|c|c|c|c|}
\hline 
Feature Set & Precision & Recall & F1 \\ 
\hline 
$\Phi_{0}$ & 72.27\% & 65.84\% & 68.91\% \\ 
\hline 
$\Phi_{1}$ & \textbf{76.49\%} & \textbf{69.65\%} &  \textbf{72.91\%} \\ 
\hline 
$\Phi_{2}$ & 72.26\% & 65.87\% & 68.92\% \\ 
\hline 
$\Phi_{3}$ & 72.35\% & 65.86\% & 68.95\% \\ 
\hline 
\end{tabular} 
\caption{Accuracy of feature sets in Table~\ref{tab:fs-1}}\label{tab:per4-1}
\end{table}

We notice that amongst the three features, function tag is the most
important feature which increases the accuracy of the baseline feature
set by about 4\% of $F_1$ score. The distance feature also helps
increase slightly the accuracy. We thus consider the fourth feature
set $\Phi_4$ defined as
\begin{equation*}
  \Phi_4 = \Phi_0 \cup \{\text{Function  Tag}\} \cup \{\text{Distance}\}.
\end{equation*}

In the fifth experiment, we modify the feature set $\Phi_4$ by
replacing the predicate with its cluster and similarly, replacing the
head word with its cluster, replacing the full path with its partial
path, resulting in feature sets $\Phi_5$, $\Phi_6$, and $\Phi_7$
respectively (see Table~\ref{tab:fs-2}). The
accuracy of these feature sets are shown in Table~\ref{tab:per4-2}.

\begin{table}[h!]
\centering
\begin{tabular}{|c|l|}
\hline 
Feature Set & Description\\ 
\hline 
$\Phi_{5}$ & $\Phi_{4} \setminus $\{Predicate\} $\cup$ \{Predicate Cluster\} \\ 
\hline 
$\Phi_{6}$ & $\Phi_{4} \setminus$\{Head Word\} $\cup$ \{Head Word Cluster\} \\ 
\hline
$\Phi_{7}$ & $\Phi_{4} \setminus$\{Full Path\} $\cup$ \{Partial Path\} \\  
\hline  
\end{tabular}
\caption{Feature sets (continued)}\label{tab:fs-2}
\end{table} 

\begin{table}[h!]
\centering
\begin{tabular}{|c|c|c|c|}
\hline 
Feature Set & Precision & Recall & F1 \\ 
\hline 
$\Phi_{4}$ & 76.60\% & 69.72\% & 73.00\% \\ 
\hline 
$\Phi_{5}$ & \textbf{76.86\%} & \textbf{70.36\%} &  \textbf{73.47\%} \\ 
\hline 
$\Phi_{6}$ & 72.50\% & 66.59\% & 69.41\% \\ 
\hline 
$\Phi_{7}$ & 76.29\% & 69.58\% & 72.78\% \\ 
\hline 
\end{tabular} 
\caption{Accuracy of feature sets in Table~\ref{tab:fs-2}}\label{tab:per4-2}
\end{table}

We observe that using the predicate cluster instead of the predicate
itself helps improve the accuracy of the system by about 0.47\% of
$F_1$ score. For ease of later presentation, we rename the feature set
$\Phi_5$ as $\Phi_8$. 

In the sixth experiment, we investigate the significance of individual
features to the system by removing them, one by one from the feature set
$\Phi_8$. By doing this, we can evaluate the importance of each
feature to our overall system. The feature sets and their corresponding
accuracy are presented in Table~\ref{tab:fs-3} and
Table~\ref{tab:per4-3} respectively.

\begin{table}
\centering
\begin{tabular}{|c|l|}
\hline 
Feature Set & Description \\ 
\hline 
$\Phi_{9}$ & $\Phi_{8} \setminus$\{Function Tag\} \\ 
\hline 
$\Phi_{10}$ & $\Phi_{8}\setminus$\{Predicate Cluster\} \\ 
\hline 
$\Phi_{11}$ & $\Phi_{8}\setminus$\{Head Word\} \\ 
\hline 
$\Phi_{12}$ & $\Phi_{8}\setminus$\{Path\} \\ 
\hline
$\Phi_{13}$ & $\Phi_{8}\setminus$\{Position\} \\ 
\hline 
$\Phi_{14}$ & $\Phi_{8}\setminus$\{Voice\} \\ 
\hline 
$\Phi_{15}$ & $\Phi_{8}\setminus$\{Subcategorization\} \\ 
\hline
$\Phi_{16}$ & $\Phi_{10} \cap \Phi_{15}$ \\
\hline    
\end{tabular} 
\caption{Feature sets (continued)} \label{tab:fs-3}
\end{table}

\begin{table}[h!]
\centering
\label{my-label}
\begin{tabular}{|c|c|c|c|}

\hline 
Feature Set & Precision & Recall & F1 \\ 
\hline 
$\Phi_{8}$ & 76.86\% & 70.36\% &  73.47\% \\ 
\hline 
$\Phi_{9}$ & 72.27\% & 66.12\% &  69.06\% \\ 
\hline
$\Phi_{10}$ & \textbf{76.87\%} & \textbf{70.41\%} &  \textbf{73.50\%} \\ 
\hline 
$\Phi_{11}$ & 72.91\% & 67.05\% & 69.86\% \\ 
\hline 
$\Phi_{12}$ & 76.81\% & 70.36\% & 73.44\% \\ 
\hline 
$\Phi_{13}$ & 76.41\% & 70.21\% & 73.18\% \\ 
\hline
$\Phi_{14}$ & 76.85\% & 70.36\% & 73.46\% \\ 
\hline  
$\Phi_{15}$ & \textbf{76.83\%} & \textbf{70.51\%} & \textbf{73.53\%} \\ 
\hline
$\Phi_{16}$ & 76.70\% & 70.31\% & 73.36\% \\ 
\hline 
\end{tabular} 
\caption{Accuracy of feature sets in Table~\ref{tab:fs-3}}\label{tab:per4-3}
\end{table}

We see that the accuracy increases slightly when either the predicate
cluster feature ($\Phi_{10}$) or the subcategorization feature
($\Phi_{15}$) is removed. However, removing both of the two features
($\Phi_{16}$) makes the accuracy decrease. For this reason, we remove
only the subcategorization feature. The best feature set includes the
following features: predicate cluster, phrase type, function tag,
parse tree path, distance, voice, position and head word. The best
accuracy of our system is 73.53\% of $F_1$ score.

\subsubsection{Learning Curve}

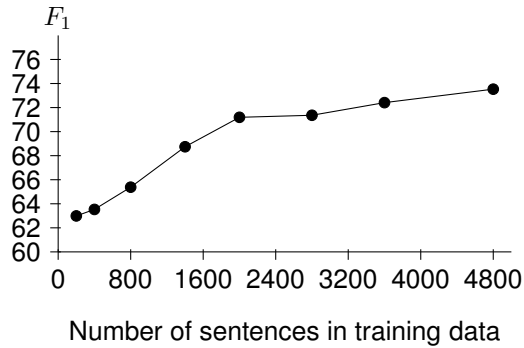
\begin{figure}[h]
\begin{center}
\begin{tikzpicture}[y=.16cm, x=.0012cm,font=\sffamily]
	\draw (0,60) -- coordinate (x axis mid) (5000,60);
    	\draw (0,60) -- coordinate (y axis mid) (0,78);
    	\foreach \x in {0,800,...,5000}
     		\draw[yshift=9.6cm] (\x,1pt) -- (\x,-3pt)
			node[anchor=north] {\x};
    	\foreach \y in {60,62,...,76}
     		\draw (1pt,\y) -- (-3pt,\y) 
     			node[anchor=east] {\y}; 
	\node[below=0.8cm] at (x axis mid) {Number of sentences in training data};
	\node[rotate=0, above=1.4cm] at (y axis mid) {$F_1$};
	
	\draw plot[ mark=*, mark options={fill=black} ]
		file {data.data};
\end{tikzpicture}
\end{center}
\caption{Learning Curve}\label{fig:lc}
\end{figure}

In the last experiment, we investigate the dependence of accuracy to
the size of the training dataset. Figure~\ref{fig:lc} depicts the
learning curve of our system when the data size is varied.

It seems that the accuracy of our system improves only slightly
starting from the dataset of about 2,000 sentences. Nevertheless, the
curve has not converged, indicating that the system could achieve a
better accuracy when a larger dataset is available.

\section{Conclusion}\label{sec:conclusion}

In this paper, we have presented the first system for Vietnamese
semantic role labelling. Our system achieves a good accuracy of about
73.5\% of $F_1$ score in the Vietnamese PropBank. 

We have argued that one cannot assume a good applicability of existing
methods and tools developed for English and other Western languages
and that they may not offer a cross-language validity. For an
isolating language such as Vietnamese, techniques developed for
inflectional languages cannot be applied ``as is''. In particular, we
have developed an algorithm for extracting argument candidates which
has a better accuracy than the 1-1 node mapping algorithm. We have
proposed some novel features which are proved to be useful for
Vietnamese SRL, notably predicate clusters and function tags. Our SRL
system, including software and corpus, is available as an open source
project for free research purpose and we believe that it is a good
baseline for the development and comparison of future Vietnamese SRL
systems.

In the future, we plan to improve further our system, in the one hand, by enlarging our
corpus so as to provide more data for the system. On the other hand,
we would like to investigate different models used in SRL, for
example joint models~\cite{Haghighi:2005} and recent inference techniques, such as integer
linear programming~\cite{Tackstrom:2015,Punyakanok:2004}.

\section*{Acknowledgements}
This work was supported by Vietnam National Foundation for Science and
Technology Development (NAFOSTED Project No. 102.05-2014.28). We would also
like to thank the FPT Technology Research Institute for providing us
the corpora for use in the experiments.

\bibliographystyle{IEEEtran} 

\bibliography{IEEEabrv,sigproc}

%
%
\end{document}